\documentclass[letterpaper, 10 pt, conference]{ieeeconf}  

\IEEEoverridecommandlockouts

\usepackage{amsmath} 
\usepackage{amssymb} 
\usepackage{multicol}
\usepackage{amsmath}
\usepackage{bm}
\usepackage{subcaption}
\usepackage{graphics} 
\usepackage{epsfig} 
\usepackage{multirow}
\usepackage{booktabs,color}
\usepackage{colortbl}
\usepackage{verbatim}
\usepackage{hyperref}
\usepackage[x11names, table]{xcolor}
\usepackage{pdfpages}

\newcommand{\greyrule}{\arrayrulecolor{black!30}\midrule\arrayrulecolor{black}}

\title{\LARGE \bf
DM-VIO: Delayed Marginalization Visual-Inertial Odometry
}

\author{Lukas von Stumberg$^{1}$ and Daniel Cremers$^{1}$
\thanks{$^{1}$All authors are with the Computer Vision Group, Technical University of Munich, Germany. {\tt\small lukas.stumberg@tum.de, cremers@in.tum.de}}%
}

\begin{document}

\thispagestyle{empty} 
\onecolumn 
 
\begin{center} 
\noindent 
 
This paper has been accepted for publication in \emph{IEEE Robotics and Automation Letters}. 

\vspace{2em} 

DOI: \href{https://doi.org/10.1109/LRA.2021.3140129}{\textcolor{blue}{10.1109/LRA.2021.3140129}}\\
 
\end{center} 
\vspace{3em} 
 
\copyright2021 IEEE. Personal use of this material is permitted. Permission from IEEE must be obtained for all other uses, in any current or future media, including reprinting/republishing this material for advertising or promotional purposes, creating new collective works, for resale or redistribution to servers or lists, or reuse of any copyrighted component of this work in other works.
 
\twocolumn

\setcounter{page}{1}

\maketitle
\thispagestyle{empty}
\pagestyle{empty}

\begin{abstract}
We present DM-VIO, a 
monocular visual-inertial odometry system based on two novel techniques called delayed marginalization and pose graph bundle adjustment.
DM-VIO performs photometric bundle adjustment with a dynamic weight for visual residuals.
We adopt marginalization, which is a popular strategy to keep the update time constrained, but it cannot easily be reversed, and linearization points of connected variables have to be fixed.
To overcome this we propose delayed marginalization:
The idea is to maintain a second factor graph, where marginalization is delayed. This allows us to later readvance this delayed graph, yielding an updated marginalization prior with new and consistent linearization points.
In addition, delayed marginalization enables us to inject IMU information into already marginalized states. This is the foundation of the proposed pose graph bundle adjustment, which we use for IMU initialization. In contrast to prior works on IMU initialization, it is able to capture the full photometric uncertainty, improving the scale estimation.
In order to cope with initially unobservable scale, we continue to optimize scale and gravity direction in the main system after IMU initialization is complete.
We evaluate our system on the EuRoC, TUM-VI, and 4Seasons datasets, which comprise flying drone, large-scale handheld, and automotive scenarios. Thanks to the proposed IMU initialization, our system exceeds the state of the art in visual-inertial odometry, even outperforming stereo-inertial methods while using only a single camera and IMU.
The code will be published at \href{http://vision.in.tum.de/dm-vio}{\nolinkurl{vision.in.tum.de/dm-vio}}
\end{abstract}

\section{Introduction}

Visual-(inertial) odometry is an increasingly relevant task with applications in robotics, autonomous driving, and augmented reality.
A combination of cameras and inertial measurement units (IMUs) for this task is a popular and sensible choice, as they are complementary sensors, resulting in a highly accurate and robust system~\cite{msckf}.
In the minimal configuration of a single camera, the IMU can also be used to recover the metric scale.
However, the scale is not always observable, the most common degenerate case being movement with a constant velocity \cite{Kaiser14imu_init}.
Hence, initialization of such system can take arbitrarily long, depending on the trajectory. 
Even worse, when initialized prematurely the IMU can in fact worsen the performance.
The difficulty of IMU initialization is why stereo-inertial methods have outperformed mono-inertial ones in the past.

Most prior systems~\cite{vinsmono} \cite{orbslamvi} \cite{orbslam3} initially run visual-only odometry and an IMU initialization in parallel. Once finished, the visual-inertial system is started. This introduces a trade-off for the duration of the initialization period: It should be as short as possible, as no IMU information is used in the main system in the meantime. But when too short, the scale estimate will be inaccurate, leading to bad performance.

VI-DSO~\cite{vidso} instead initializes immediately with an arbitrary scale, and explicitly optimizes the scale in the main system.
This yields highly accurate scale estimates, but it can significantly increase the time until the scale is correctly estimated. Also, it can fail in cases where the initial scale error is very high, like in large-scale outdoor environments.

We propose a combination of the two strategies: Similar to the former, we start with a visual-only system and run an IMU initializer in parallel. But after IMU initialization we still estimate scale and gravity direction as explicit optimization variables in the main system. This results in a quickly converging and highly accurate system.

\begin{figure}[t]
    \centering
    \includegraphics[width=\linewidth]{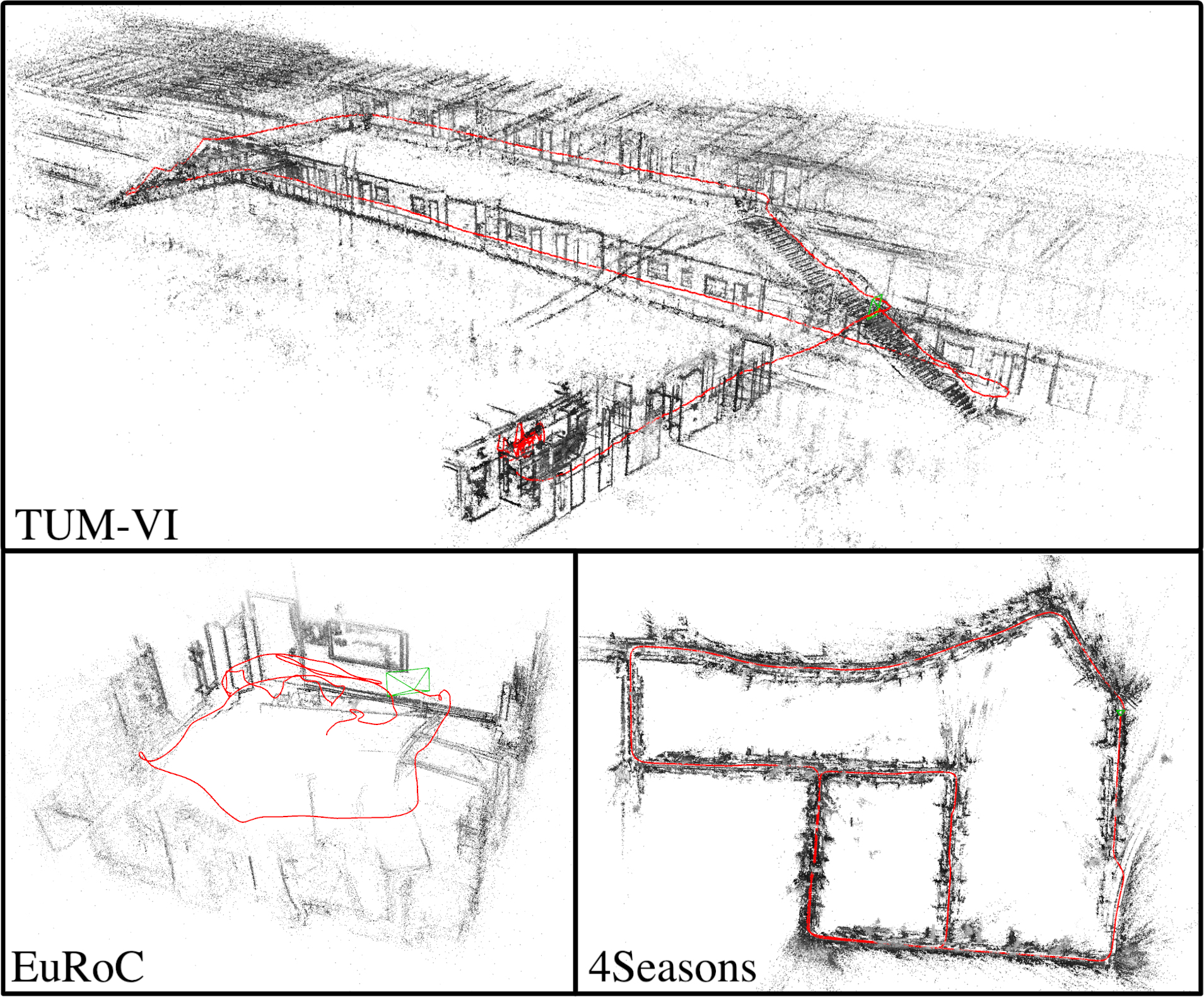}
    \caption{In this paper, we propose a novel method for monocular visual-inertial odometry. It provides state-of-the-art performance on three different benchmarks.
    Here we show pointclouds and trajectories (red) for magistrale5, V203\_difficult, and neighbor\_2020-03-26\_13-32-55\_0.}
    \label{fig:teaser} 
\end{figure}

This initialization strategy can lead to three questions: 
1) How can the visual uncertainty be properly captured in the IMU initializer. 2) How can information about scale and IMU variables be transferred from the IMU initializer to the main system? 3) If the scale estimate changes, how can a consistent marginalization prior be maintained?
VI-DSO~\cite{vidso} tried to address 3 by introducing dynamic marginalization, which does keep the marginalization factor consistent, but loses too much information in the process.

\emph{In this work we propose delayed marginalization, which provides a meaningful answer to all three of these questions}.
The idea is to maintain a second, \emph{delayed} marginalization prior, which has very little overhead, but enables three core techniques:
\begin{enumerate}
    \item We can populate the delayed factor graph with new IMU factors to perform the proposed pose graph bundle adjustment (PGBA). This is the basis of an IMU initialization which captures the full photometric uncertainty, leading to increased accuracy.
    \item The graph used for IMU initialization can be \emph{re-advanced}, providing a marginalization prior with IMU information for the main system.
    \item When the scale changes significantly in the main system we can trigger \emph{marginalization replacement}. 
\end{enumerate}

The combination of these techniques makes for a highly accurate initializer, which is robust even to long periods of unobservability.
Based on it we implement a visual-inertial odometry (VIO) system featuring a photometric front-end integrated with a new dynamic photometric weight.

We evaluate our method on three challenging datasets (Fig.~\ref{fig:teaser}), capturing three domains: 
The EuRoC dataset~\cite{euroc} recorded by a flying drone, the TUM-VI dataset~\cite{tumvi} captured with a handheld device, and the 4Seasons dataset~\cite{4seasons} representing the automotive scenario.
The latter features long stretches of constant velocity, posing a particular challenge for mono-inertial odometry.

We show that our system exceeds the state of the art in visual-inertial odometry, even outperforming stereo-inertial methods.
In summary our contributions are:
\begin{itemize}
    \item Delayed marginalization compensates drawbacks of marginalization while retaining the advantages.
    \item Pose graph bundle adjustment (PGBA) combines the efficiency of pose graph optimization with the full uncertainty of bundle adjustment.
    \item A state-of-the-art visual-inertial odometry system with a novel multi-stage IMU initializer and dynamically weighted photometric factors.
\end{itemize}
The full source code for our approach will be released. 

\section{Related Work}
Initially, most visual odometry and SLAM systems have been feature-based \cite{nister2004_vo}, either using filtering \cite{davison07pami} or nonlinear optimization \cite{ptam} \cite{orbslam}. More recently, direct methods have been proposed, which optimize a photometric error function and can operate on dense \cite{kerl13icra} \cite{dtam}, semi-dense~\cite{lsdslam}, or sparse point clouds~\cite{dso}. 

Mourikis and Roumeliotis \cite{msckf} have shown that a tight integration of visual and inertial measurements can greatly increase accuracy and robustness of odometry. Afterwards, many tightly-coupled visual-inertial odometry \cite{rovio} \cite{okvis} and SLAM systems \cite{basalt} \cite{kimera} \cite{vinsmono} \cite{orbslam3} have been proposed.

    Initialization of monocular visual-inertial systems is not trivial, as sufficient motion is necessary for the scale to become observable \cite{Martinelli2014} \cite{Kaiser14imu_init}. Most systems \cite{orbslamvi} \cite{vinsmono} \cite{orbslam3} start with a visual-only system and use its output for a separate IMU initialization.  In contrast to these systems, we continue optimizing the scale explicitly in the main system. 
    We note that ORB-SLAM3~\cite{orbslam3} also continues to refine the scale after initialization, but this is a separate optimization fixing all poses and only performed until 75 seconds after initialization.
    \cite{onlinescale} also continues to optimize the scale in the main system, but in contrast to us they do not transfer covariances between the main system and the initializer, thus they do not achieve the same level of accuracy.
    Different to all these systems, the proposed delayed marginalization allows our IMU initializer to capture the full visual uncertainty and continuously optimize the scale in the main system.

VI-DSO \cite{vidso} initializes immediately with an arbitrary scale and explicitly optimizes the scale in the main system. It also introduced dynamic marginalization to handle the consequential large scale changes in the main system. Compared to it we propose a separate IMU initializer, delayed marginalization as a better alternative to dynamic marginalization, a dynamic photometric error weight, and more improvements, resulting in greatly improved accuracy and robustness.

\section{Method}

\subsection{Notation} 
We denote vectors as bold lowercase letters $\mathbf{x}$, matrices as bold upper-case letter $\mathbf{H}$, scalars as lowercase letters $\lambda$, and functions as uppercase letters $E$.
$\mathbf{T}_{\text{w}\_\text{cam}_i}^V \in \mathbf{SE}(3)$ represents the transformation from camera $i$ to world in the visual coordinate frame $V$, and $\mathbf{R}_{\text{w}\_\text{cam}_i}^V \in \mathbf{SO}(3)$ is the respective rotation.
Poses are represented either in visual frame $\mathbf{P}_i^V := \mathbf{T}_{\text{cam}_i\_\text{w}}^V$, or in inertial frame $\mathbf{P}_i^I := \mathbf{T}_{\text{w}\_\text{imu}_i}^I$. If not mentioned otherwise we use poses in visual frame $\mathbf{P}_i := \mathbf{P}_i^V$. 
We also use states $\mathbf{s}$, which can contain transformations, rotations, and vectors. For states we define the subtraction operator $\mathbf{s}_i \boxminus \mathbf{s}_j$, which applies $\log(\mathbf{R}_i \mathbf{R}_j^{-1})$ for rotations and other Lie group elements, and a regular subtraction for vector values.

\subsection{Direct Visual-Inertial Bundle Adjustment}\label{sec:ba}
The core of DM-VIO is the visual-inertial bundle adjustment performed for all keyframes.
As commonly done, we jointly optimize visual and IMU variables in a combined energy function. For the visual part we choose a direct formulation based on DSO \cite{dso}, as it is a very accurate and robust system. For integrating IMU data into the bundle adjustment we perform preintegration \cite{preintegration-forster} between keyframes.

We optimize the following energy function using the Levenberg-Marquardt algorithm: 

\begin{equation}
    E(\mathbf{s}) = W(e_\text{photo}) \cdot E_\text{photo} + E_\text{imu} + E_\text{prior}
\end{equation}
$E_\text{prior}$ contains added priors on the first pose and the gravity direction, as well as the marginalization priors explained in section~\ref{sec:marginalization}.
In the following we describe the individual energy terms and the optimized state.

\noindent\textbf{Photometric error: } 
The photometric energy is based on \cite{dso}. 
We optimize a set of active keyframes $\mathcal{F}$, each of which hosts a set of points $\mathcal{P}_i$. Every point $\mathbf{p}$ is projected into all keyframes $\text{obs}(\mathbf{p})$ where it is visible, and the photometric energy is computed:
\begin{equation}
E_{\text{photo}} = \sum_{i \in \mathcal{F}} \sum_{\mathbf{p} \in \mathcal{P}_i} \sum_{j \in \text{obs}(\mathbf{p})} E_{\mathbf{p}j}
\end{equation}
\begin{equation}\label{eq:photometric}
E_{\mathbf{p}j} = \sum_{\mathbf{p} \in \mathcal{N}_{\mathbf{p}}} \omega_{\mathbf{p}} \bigg \lVert (I_j[\mathbf{p'}] - b_j) - \frac{t_j e^{a_j}}{t_i e^{a_i}} (I_i[\mathbf{p}] - b_i) \bigg \rVert_{\gamma}
\end{equation}
For details regarding the variables 
we refer the reader to \cite{dso}.

\noindent\textbf{Dynamic photometric weight: } 
In cases of bad image quality, 
the system should rely mostly on the inertial data. However due to the photometric cost function used, bad image quality will often lead to very large photometric residuals, effectively increasing the photometric weight compared to the IMU.
To counteract this we propose a dynamic photometric weight $W(e_\text{photo})$. We compute it using the root mean squared photometric error $e_\text{photo} = \sqrt{E_\text{photo} / n_\text{residuals}}$.
\begin{equation}
    W(e_\text{photo}) = \lambda \cdot 
    \begin{cases} 
        (\theta / e_\text{photo})^2, &\text{if } e_\text{photo} \geq \theta \\
        1, &\text{otherwise}
    \end{cases}
\end{equation}
where $\lambda$ is a static weight component, and $\theta$ is the threshold from which the error-dependent weight is activated.
This effectively normalizes the root mean squared photometric error to be $\sqrt{\lambda}\theta$ at maximum, similar to a threshold robust cost function~\cite{atallcosts}. In contrast to the Huber norm in Equation (\ref{eq:photometric}), which downweights individual points that violate the photometric assumption, this weight addresses cases where the overall image quality is bad and increases the relative weight of the IMU. 
In our experiments we choose $\theta=8$.

\noindent\textbf{Optimized variables: }
We optimize scale and gravity direction as explicit variables. While bundle adjustment can in principle also change the scale and global orientation, convergence is improved when optimizing them explicitly instead \cite{vidso}. 
To facilitate this, we represent poses for the visual factors in visual frame $V$ and poses for the IMU factors in IMU frame $I$. Whereas the IMU frame has a metric scale and a z-axis aligned with gravity direction, the visual frame can have an arbitrary scale and rotation, which is defined during initialization of the visual system. To model this we optimize the scale $s$ and the rotation $\mathbf{R}_{V\_I}$.
As yaw is not observable using an IMU, we fix the last coordinate of $\mathbf{R}_{V\_I}$.
We convert between the coordinate frames using:
\begin{equation} \label{eq:omega}
    \begin{split}
        \mathbf{P}_i^I := \mathbf{T}_{\text{w}\_\text{imu}_i}^I = \Omega(\mathbf{P}_i^V, \mathbf{S}, \mathbf{R}_{V\_I}) =  \\
        \mathbf{R}_{V\_I}^{-1} \mathbf{S}_{I\_V} (\mathbf{P}_i^V)^{-1} \mathbf{S}_{I\_V}^{-1} \mathbf{T}_{\text{cam}\_\text{imu}}
    \end{split}
\end{equation}
where $\mathbf{S}_{I\_V}$ is the $\mathbf{Sim}(3)$ element with identity rotation and translation, and scale $s$. 
The other variables are converted to $\mathbf{Sim}(3)$, but note that the result has scale 1 and is in $\mathbf{SE}(3)$.

The full state optimized is 
\begin{equation}
    \mathbf{s}= \{ s, \mathbf{R}_{V\_I} \} \cup \bigcup\limits_{i \in \mathcal{F}} \mathbf{s}_i
\end{equation}
with $\mathbf{s}_i$ being the states for all active keyframes defined as:
\begin{equation}
    \mathbf{s}_i = \{ \mathbf{P}_i^V, \mathbf{v}_i, \mathbf{b}_i, a_i, b_i, d_i^0, d_i^2, ... d_i^j \}
\end{equation} 
where $\mathbf{v}_i$ is the velocity, $\mathbf{b}_i$ the bias, $a_i$ and $b_i$ are affine brightness parameters, and $d_i^j$ are the inverse depths of active points hosted in the keyframe.
Optimization is performed with a custom integration of the SIMD-accelerated code from \cite{dso} for photometric residuals and GTSAM for other factors.

\noindent\textbf{IMU Error: } We apply the well-known IMU preintegration first proposed in~\cite{preintegration1}, implemented as smart factors in~\cite{preintegration2}, and further improved in~\cite{preintegration-forster}. 
For this energy we use the IMU state $\mathbf{s}_i^I := \{ \mathbf{P}_i^I, \mathbf{v}_i, \mathbf{b}_i \}$,
which contains poses in IMU frame and is computed from the optimized state $\mathbf{s}_i$ using Equation (\ref{eq:omega}).
Given the previous state $\mathbf{s}_i^I$, the preintegration data provides us with a prediction $\widehat{\mathbf{s}}_j^I$ for the following state $\mathbf{s}_j^I$ as well as a covariance matrix $\widehat{\boldsymbol{\Sigma}}_j$. The resulting inertial error function penalizes deviations of the current state estimate from the predicted state.
\begin{equation}
E_{\text{imu}}( \mathbf{s}_i^I, \mathbf{s}_j^I ) := \left(\widehat{\mathbf{s}}_j^I \boxminus \mathbf{s}_j^I \right)^T \widehat{\boldsymbol{\Sigma}}_j^{-1} \left(\widehat{\mathbf{s}}_j^I \boxminus \mathbf{s}_j^I \right)
\end{equation}

\subsection{Partial Marginalization using the Schur Complement} \label{sec:marginalization}
We marginalize old variables using the Schur complement. 
When marginalizing a set $\beta$ of variables, we gather all factors dependent on them as well as the connected variables $\alpha$, which form the Markov blanket. These factors are linearized at the current state estimate, yielding the linear system: 
\begin{equation}
 \begin{bmatrix}
     \mathbf{H}_{\alpha\alpha} & \mathbf{H}_{\alpha\beta}  \\
     \mathbf{H}_{\beta\alpha} & \mathbf{H}_{\beta\beta}
\end{bmatrix}
\begin{bmatrix}
    \mathbf{s}_\alpha \\
    \mathbf{s}_\beta
\end{bmatrix} = 
\begin{bmatrix}
    \mathbf{b}_\alpha \\
    \mathbf{b}_\beta
\end{bmatrix}
\end{equation}
We apply the Schur-complement, which results in the new linear system
$\widehat{\mathbf{H}_{\alpha\alpha}} \bm{s}_\alpha = \widehat{\bm{b}_\alpha}$ with
\begin{equation}
    \widehat{\mathbf{H}_{\alpha\alpha}} = \mathbf{H}_{\alpha\alpha} - \mathbf{H}_{\alpha\beta} \mathbf{H}_{\beta\beta}^{-1} \mathbf{H}_{\beta\alpha}
\end{equation}
\begin{equation}
    \widehat{\bm{b}_\alpha} = \bm{b}_\alpha - \mathbf{H}_{\alpha\beta} \mathbf{H}_{\beta\beta}^{-1} \bm{b}_\beta
\end{equation}
This linear system forms a marginalization prior connecting all variables in $\alpha$ (Fig.~\ref{fig:delayedmarg}a).

We keep a maximum of $N_f=8$ keyframes during the bundle adjustment\footnote{We define $N_f$ as the maximum number of frames during bundle adjustment, whereas in \cite{dso} it is the number of frames after marginalization.}. The marginalization strategy is taken over from \cite{dso}: This means that different from a fixed-lag smoother, we do not always marginalize the oldest pose, but instead keep a combination of newer and older poses, as long as they do not leave the field of view. As shown in \cite{dso} this is superior to a fixed-lag smoother for visual odometry. 
When marginalizing a pose, first all remaining points hosted in the frame are marginalized and residuals with remaining active points are dropped. This retains sparsity of the Hessian while preserving enough information.

\subsection{Delayed Marginalization}\label{sec:delayedmarg}

\begin{figure*}[htb]
    \centering
    \includegraphics[width=\linewidth]{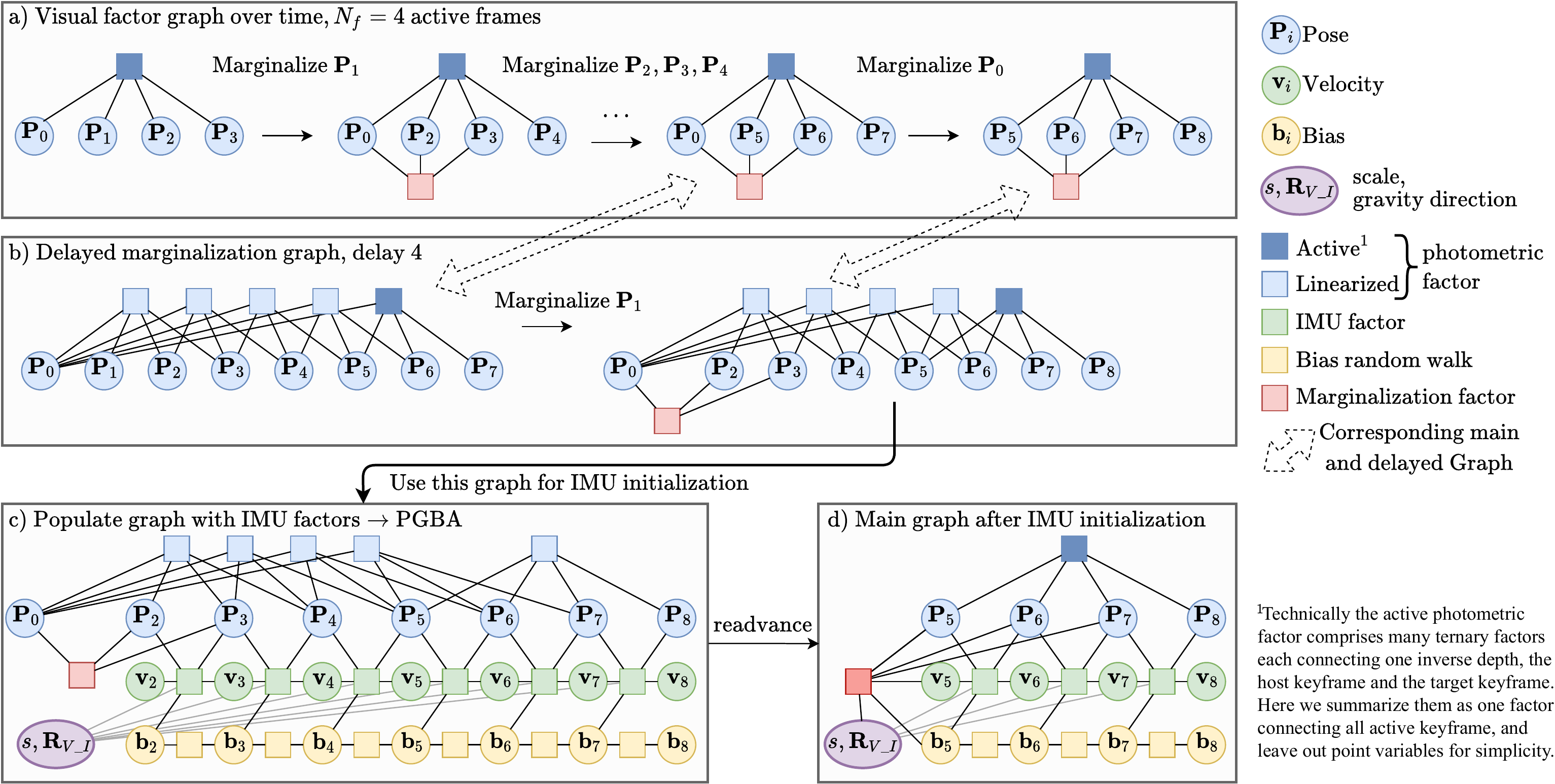} 
    \caption{Delayed Marginalization and PGBA: a) Normal marginalization in the visual graph. Note that not always the oldest pose is marginalized. b) Delayed marginalization: We marginalize all variables in the same order as the main graph, but with a delay $d$ (in practice $d=100$). Marginalization in this graph is equally fast as marginalizing in the main graph. c)~For the pose graph bundle adjustment (PGBA) we populate the delayed graph with IMU factors. This optimization leverages the full photometric uncertainty. d) We readvance the marginalization in the graph used for PGBA to obtain an updated marginalization prior for the main system. This transfers inertial information from the initializer to the main system.
    }
    \label{fig:delayedmarg} 
\end{figure*}

The concept of marginalization explained in the previous section has the advantage of capturing the full probability distribution. In fact, solving the resulting smaller system is equivalent to solving the much larger original system, as long as the marginalized factors are not relinearized.

However, it also comes with severe drawbacks: Reverting the marginalization of a set of variables is not possible without redoing the whole marginalization procedure. Also, to keep the marginalization prior consistent, 
First-Estimates Jacobians (FEJ)~\cite{fej} have to be applied. This means that the linearization point of all connected variables has to be fixed as soon as they are connected to a marginalization prior.
This is especially problematic for visual-inertial odometry, where the scale is connected to the marginalization prior as soon as the first keyframe is marginalized, but might change significantly.
In~\cite{vidso} dynamic marginalization was introduced to combat this, but it is limited in its application to a single one-dimensional variable, namely the scale, and loses most prior inertial information when the scale changes quickly.

Here we introduce delayed marginalization which circumvents the drawbacks of marginalization while retaining the advantages. It enables us to:
\begin{itemize}
    \item Effectively undo part of the marginalization to capture the full photometric probability distribution for the pose graph bundle adjustment (section \ref{sec:pgba}). 
    \item Update the initially visual-only marginalization prior with IMU information after the IMU initialization.
    \item Relinearize variables in the Markov blanket while keeping all visual and most inertial information.
\end{itemize}

\noindent{\textbf{The idea of delayed marginalization is that marginalization cannot be undone, but it can be delayed:} \emph{In addition} to the normal marginalization prior we also maintain a second, \emph{delayed} marginalization prior and corresponding factor graph. 
In this delayed graph, marginalization of frames is performed with a delay of $d$. 
Points are still marginalized at the same time in the delayed graph, resulting in linearized photometric factors.
We note that the same marginalization order as in the original graph is preserved. 
Switching to a fixed-lag smoother for this graph would immediately lead to a much larger Markov blanket jeopardizing the runtime of the system.
E.g. in Fig.~\ref{fig:delayedmarg}b we depict the delayed marginalization of $\mathbf{P}_1$. The Markov blanket only contains $\mathbf{P}_0$, $\mathbf{P}_2$, and $\mathbf{P}_3$. If we instead marginalized the oldest frame $\mathbf{P}_0$, the Markov blanket would contain $\mathbf{P}_1 - \mathbf{P}_7$, leading to higher runtime. 

\noindent\textbf{Marginalization in the delayed graph has the same runtime as marginalization in the original graph.} The delayed graph contains the same photometric factors as the original graph, and points are marginalized at the same time. 
This means that each linearized photometric factor in the delayed graph is connected to exactly the $N_f=8$ keyframes which were active when the respective factor was generated. By keeping the marginalization order, the Markov blanket in the delayed graph always has the same size as the one in the original graph. Thus, the runtime of the Schur complement is the same. This means that the overhead of Delayed Marginalization is very small even for arbitrarily large delays, as it only amounts to an additional marginalization procedure per delayed graph.

\subsection{Pose Graph Bundle Adjustment for IMU Initialization} \label{sec:pgba}
PGBA utilizes delayed marginalization for IMU initialization.
The idea is to populate the delayed graph with IMU factors and optimize all variables
(Fig.~\ref{fig:delayedmarg}c). 

\noindent\textbf{Populating the graph: } 
    Let a frame $\mathbf{P}_i$ be directly connected to the newest pose $\mathbf{P}_k$ iff all poses $\mathbf{P}_j, i < j < k$ have not been marginalized yet.
    We determine the first frame $\mathbf{P}_\text{conn}$ in the delayed graph which is still directly connected to the newest frame.    In Fig.~\ref{fig:delayedmarg}c this is $\mathbf{P}_2$. From there, we insert IMU factors and bias factors to all successive frames. 

We cannot start before $\mathbf{P}_\text{conn}$ because we do not want to insert IMU factors between non-successive keyframes.
As the marginalization order is not fixed-lag, this means that we have to optimize poses 
without corresponding IMU variables. 

\textbf{It can be shown that there can be at most $N_f-2$ poses without IMU variables.} The reason is that all non-connected poses were at some point active at the same time.
This means that in practice we have at least $d - N_f + 2$ poses for which we can add IMU data. 
In practice, we choose $N_f=8$ and delay $d=100$, meaning that even in the worst case there will be $93$ IMU factors in the optimization.
As explained previously, fixed-lag smoothing would either result in a dense Hessian or in suboptimal performance of the visual system, so this is a very good trade-off.

\noindent\textbf{Optimization: }  We optimize the graph with the GTSAM \cite{gtsam} library using the Levenberg-Marquardt optimizer with the provided Ceres-default settings.
In this optimization all points are marginalized. 
We call it pose graph bundle adjustment because it is a combination of regular pose graph optimization (PGO) and bundle adjustment (BA).
In contrast to BA, we do not update our estimates for point depths and do not relinearize photometric error terms. Different from PGO we do not use binary constraints between poses, but instead use "octonary" constraints, which connect $N_f$
frames and capture the full probability distribution of BA. Compared to PGO our solution is thus more accurate while being much faster than full BA. By using a fixed delay it is also constrained in runtime even though it can be performed at any time without losing any prior visual information.

\noindent\textbf{Readvancing: } Another advantage of delayed marginalization and our PGBA is that we can obtain a marginalization prior for the main system, capturing all the visual and inertial information. For this, we readvance the graph used for the PGBA. 
This works by successively marginalizing all the variables which have been marginalized in the main graph. Again, this is done preserving the marginalization order, which means that in each marginalization step the Markov blanket has a fixed maximum size. Hence, marginalizing step by step is significantly faster than marginalizing all variables at once, which would involve a much larger matrix inversion.
Fig.~\ref{fig:delayedmarg}d shows the result of readvancing.

\subsection{Robust Multi-Stage IMU Initialization}
\begin{figure}[htb]
    \centering
    \includegraphics[width=0.95\linewidth]{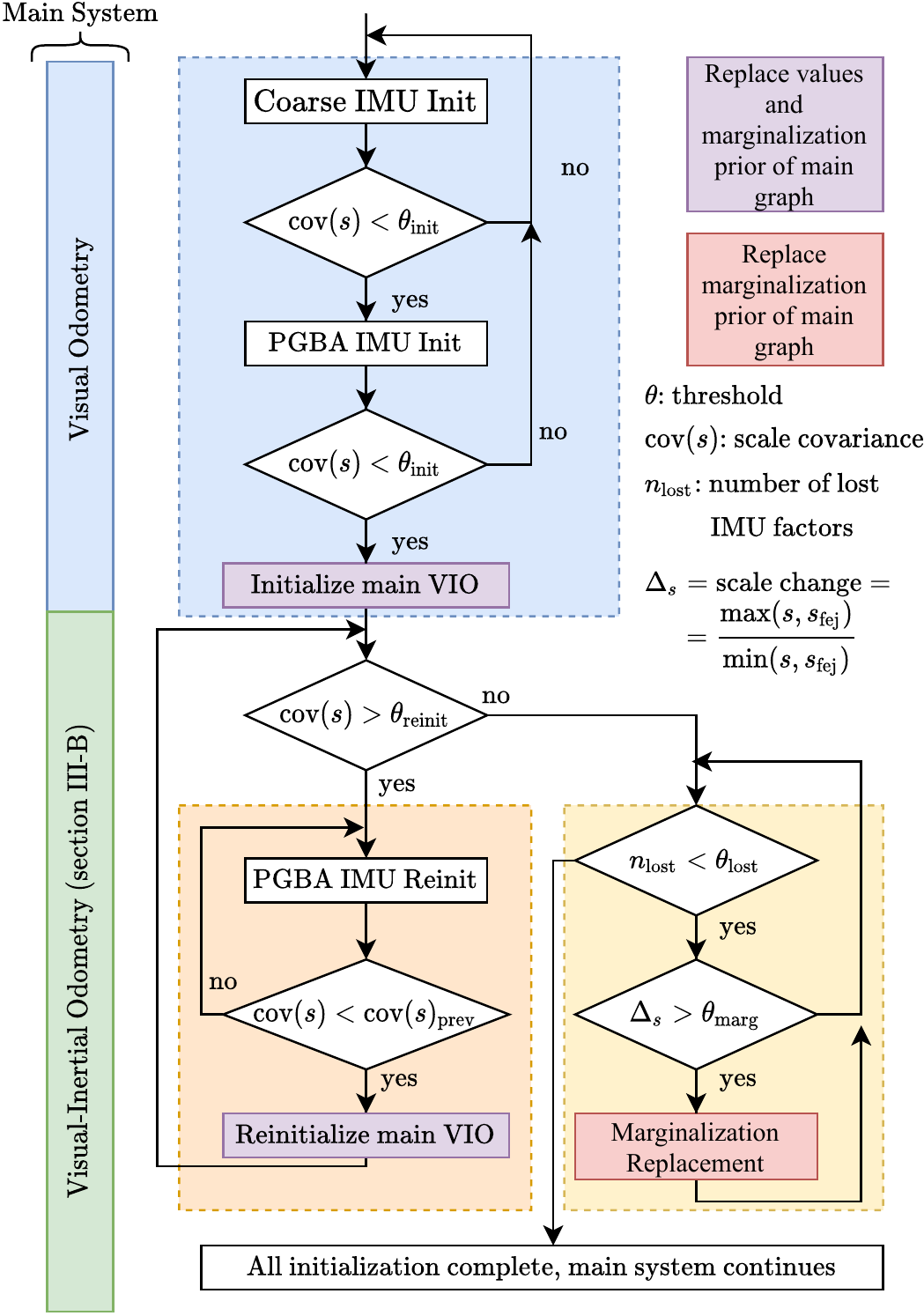} 
    \caption{
    Our multi-stage IMU initialization. First we perform a coarse IMU initialization, which provides initial values for the PGBA. The PGBA captures the full visual covariances, achieving very accurate initial estimates for scale, gravity direction, and biases. 
    It also provides an updated marginalization prior for the main graph. 
By also optimizing the scale in the main VIO system (green box),
we can initialize early (purple box) and later reinitialize or perform marginalization replacement, if new information about the scale becomes available. The proposed delayed marginalization is what enables both, the PGBA, and the marginalization replacement.
} 
    \label{fig:initstages} 
\end{figure}
Our initialization strategy is based on three insights:
\begin{enumerate}
    \item When some variables are unknown (in our case scale, gravity direction, and biases) and others are close to the optimum, it is most efficient to first optimize only the unknown variables and fix the others.
    \item The most accurate result can be obtained by optimizing all variables jointly, capturing the full covariance.
    \item When marginalizing, connected variables have to be close to the optimum, otherwise the marginalization prior becomes inconsistent.
\end{enumerate}

These observations inspire 1) the Coarse IMU Initialization, 2) the PGBA, and 3) the Marginalization Replacement (Fig.~\ref{fig:initstages}).
Note that after "Initialize main VIO", the main VIO system~\ref{sec:ba} (green box) is already running in parallel.

For this initializer we use a single delayed graph with a delay of $d=100$. This delayed graph will always contain only visual factors and no IMU factors, even after the first initialization, to facilitate the marginalization replacement.

\begin{table*}[ht]
\caption{
    Evaluation of various mono (M) and stereo (S) visual-inertial odometry systems on EuRoC. Our system provides a notable improvement over the state-of-the art. Please note that a full SLAM system utilizing loop closures can achieve even more accurate results, e.g. ORB-SLAM-VI has a mean error of 0.075, and ORB-SLAM3 has a mean error of 0.043.
}
\label{tab:euroc}
\begin{center}
\resizebox{\textwidth}{!}{%
  \begin{tabular}{c c|c c c c c c c c c c c | c}
    \toprule
Sequence &  & MH1 & MH2 & MH3 & MH4 & MH5 & V11 & V12 & V13 & V21 & V22 & V23 & Avg \\
\greyrule
MCSKF$^2$~\cite{msckf} (M)  & RMSE & 0.42 & 0.45 & 0.23 & 0.37 & 0.48 & 0.34 & 0.20 & 0.67 & 0.10 & 0.16 & 1.13 & 0.414 \\
\greyrule
OKVIS$^1$~\cite{okvis} (M) & RMSE & 0.33 & 0.37 & 0.25 & 0.27 & 0.39 & 0.094 & 0.14 & 0.21 & 0.090 & 0.17 & 0.23 & 0.231 \\ 
\greyrule
ROVIO$^2$~\cite{rovio} (M) & RMSE & 0.21 & 0.25 & 0.25 & 0.49 & 0.52 & 0.10 & 0.10 & 0.14 & 0.12 & 0.14 & 0.14 & 0.224 \\
\greyrule
VINS-Mono~\cite{vinsmono} (M) & RMSE & 0.15 & 0.15 & 0.22 & 0.32 & 0.30 & 0.079 & 0.11 & 0.18 & 0.080 & 0.16 & 0.27 & 0.184 \\ 
\greyrule
Kimera~\cite{kimera} (S) & RMSE & 0.11 & 0.10 & 0.16 & 0.24 & 0.35 & 0.05 & 0.08 & 0.07 & 0.08 & 0.10 & 0.21 & 0.141 \\
\greyrule
Online VIO~\cite{onlinescale} (M) & RMSE & 0.14 & 0.13 & 0.20 & 0.22 & 0.20 & 0.05 & 0.07 & 0.16 & 0.04 & 0.11 & 0.17 & 0.135 \\
\greyrule
\multirow{2}{*}{\begin{tabular}{c}VI-DSO~\cite{vidso} (M)\end{tabular}} & RMSE & \textbf{0.062} & \textbf{0.044} & 0.117 & 0.132 & 0.121 & 0.059 & 0.067 & 0.096 & 0.040 & 0.062 & 0.174 & 0.089 \\
   & Scale Error (\%) & 1.1 & 0.5 & 0.4 & 0.2 & 0.8 & 1.1 & 1.1 & 0.8 & 1.2 & 0.3 & 0.4 & 0.7 \\
\greyrule
BASALT~\cite{basalt} (S) & RMSE &             0.07 & 0.06 & \textbf{0.07} & 0.13 & 0.11 & \textbf{0.04} & 0.05 & 0.10 & 0.04 & \textbf{0.05} & - & 0.072 \\
\greyrule
\multirow{2}{*}{\begin{tabular}{c}DM-VIO (M)\end{tabular}} & RMSE & 0.065 & \textbf{0.044} & 0.097 & \textbf{0.102} & \textbf{0.096} & 0.048 & \textbf{0.045} & \textbf{0.069} & \textbf{0.029} & \textbf{0.050} & \textbf{0.114} & \textbf{0.069} \\
                                                                & Scale Error (\%) & 1.3 & 0.9 & 0.4 & 0.2 & 0.4 & 0.4 & 1.0 & 0.3 & 0.02 & 0.6 & 0.8 & 0.6 \\

\bottomrule

\multicolumn{14}{l}{
    \begin{tabular}{l}
        \addlinespace[2ex]
        $^1$ results taken from \cite{vinsmono}. \\
        $^2$ results taken from  \cite{benchmarkpaper}, these are $\mathbf{Sim}(3)$-aligned.\\
        All other results are taken from the respective paper.
    \end{tabular}
}

\vspace{-0.3cm}
  \end{tabular}
}
\end{center} 
\end{table*}

\noindent\textbf{Coarse IMU Initialization: } For this we only consider the last $d=100$ keyframes and connect them with IMU factors. 
Similar to the inertial only optimization used for initialization in ORB-SLAM3~\cite{orbslam3}, in this optimization we fix the poses and use a single bias.
We only optimize velocities, bias, the gravity direction and the scale.
Gravity direction is initialized by averaging the accelerometer measurements between the first two keyframes, scale is initialized with 1, and bias and velocity with 0. 
This optimization is less accurate than PGBA but serves as an initialization for it. After optimizing, we compute the marginal covariance for the scale~$\text{cov}(s)$ and continue to the PGBA if it is smaller than a threshold~$\theta_\text{init}$.
As shown in \cite{camposinit}, taking into account IMU noise parameters is crucial for good IMU initialization, which our coarse IMU initialization satisfies. But for our method it is just an initialization for the PGBA, which in addition models photometric noise properties.

\noindent\textbf{PGBA IMU Init.: } We perform PGBA as explained in section \ref{sec:pgba}. Afterwards, we again threshold on the marginal covariance for the scale to find out if the optimization was successful. When a tighter threshold $\theta_\text{reinit}$ is not also met, we initialize with the result, but will perform another PGBA afterwards to reinitialize with more accurate values. 
This reinitialization enables us to set $\theta_\text{init}$ to a relatively large value, allowing to use IMU data in the main system earlier.

\noindent\textbf{Marginalization Replacement: } After IMU initialization, we monitor how much the scale $s$ changes compared to the First-Estimates scale $s_\text{fej}$ used in the marginalization prior. If this change exceeds a threshold $\theta_\text{marg}$, i.e. $\delta_s:=\max(s, s_\text{fej}) / \min(s, s_\text{fej}) > \theta_s$, we trigger a marginalization replacement. For the marginalization replacement we rebuild the PGBA graph by populating the delayed graph with IMU factors, Fig.~\ref{fig:delayedmarg}c). Different from the PGBA, we do not optimize in this graph but instead just readvance it to obtain an updated marginalization prior. This new prior still contains all visual factors and at least the last $d - N_f + 1=93$ IMU factors. We disable the marginalization replacement if more than $\theta_\text{lost}=50\%$ of the IMU factors contained in the previous prior would be lost. 
This procedure shows how delayed marginalization can be used to update FEJ values, overcoming one of the main problems of marginalization. 

In realtime mode we perform the coarse IMU initialization and the PGBA in a separate thread.
Note how important the proposed delayed marginalization is for this IMU initialization. It allows the PGBA to capture the full covariance from the photometric bundle adjustment. By readvancing, this also enables us to generate a marginalization prior for the main system, containing all IMU information from the initializer. Lastly, it is used for updating the marginalization prior when the scale changes after the initialization.

\section{Results}
We evaluate our method on the EuRoC dataset~\cite{euroc}, the TUM-VI dataset~\cite{tumvi}, and the 4Seasons dataset~\cite{4seasons}, covering flying drones, handheld sequences, and autonomous driving respectively. 
We encourage the reader to watch the supplementary video which shows qualitative realtime results on 4Seasons and TUM-VI slides1.
We also provide ablation studies and runtime evaluations in the supplementary available at \href{http://vision.in.tum.de/dm-vio}{\nolinkurl{vision.in.tum.de/dm-vio}}.

Unless otherwise stated all experiments are performed in realtime mode on the same MacBook Pro 2013 (i7 at 2.3GHz) which was used for generating the results in~\cite{vidso}, without utilizing the GPU.
As ORB-SLAM3 is not officially supported on MacOS, we show results for it on a slightly stronger desktop with an Intel Core i7-7700K at 4.2GHz, which is very similar to the PC used in their paper. 

All methods are evaluated 10 times for EuRoC and 5 times for the other datasets on each sequence. 
Following \cite{dso}, results are presented in cumulative error plots, which show how many sequences (y-axis) have been tracked with an accuracy better than the threshold on the x-axis.
We perform $\mathbf{SE}(3)$ alignment of the trajectory with the provided ground-truth and report the root mean squared error (RMSE), also called absolute trajectory error (ATE). On TUM-VI and 4Seasons, trajectory lengths can vary greatly so we report the drift in $\%$, which we compute with $\text{drift}=\frac{\text{rmse}\cdot 100}{\text{length}}$.
We also show tables to compare to numbers from other papers and report the median result for each sequence for our method.

\subsection{EuRoC dataset}
The EuRoC dataset~\cite{euroc} is the most popular visual-inertial dataset to date, and many powerful methods have been evaluated on it. 
In Table~\ref{tab:euroc} we compare to the state-of-the art in visual-inertial odometry, all results are without loop-closure.
Our method outperforms all other methods clearly in terms of RMSE. The closest competitor is Basalt~\cite{basalt}, a stereo-inertial method which achieves a smaller error on 2 sequences. We also observe the lowest average scale error reported on the dataset so far, confirming that our contributions in IMU initialization have a positive impact on performance.
In the supplementary we provide runtime evaluations, showing that tracking takes 10.34ms on average, and keyframe processing takes 53.67ms. The delayed marginalization is responsible for an overhead of 0.44ms or 0.8\% in the keyframe thread.

\subsection{TUM-VI dataset}

\begin{table}
    \caption{RMSE ATE in m on the TUM-VI dataset~\cite{tumvi}. Best results in bold, underline is the best result among monocular methods. DM-VIO outperforms even state-of-the-art stereo-inertial methods by a large margin. 
    }
\label{tab:tumvi}
    \centering
    \resizebox{\columnwidth}{!}{%
    \begin{tabular}{l|rrrrrr}
        \toprule        
        Sequence                & ROVIO         & VINS               & OKVIS          & BASALT         & DM-VIO                 & length \\
                                & stereo        & mono               & stereo         & stereo         & mono                   & [m] \\
        \midrule                                                                                       
        corridor1               & 0.47          & 0.63               & 0.33           & 0.34           &   \textbf{0.19}        & 305 \\
        corridor2               & 0.75          & 0.95               & 0.47           & \textbf{0.42}  &   \underline{0.47}     & 322 \\
        corridor3               & 0.85          & 1.56               & 0.57           & 0.35           &   \textbf{0.24}        & 300 \\
        corridor4               & \textbf{0.13} & 0.25               & 0.26           & 0.21           &   \textbf{0.13}        & 114 \\
        corridor5               & 2.09          & 0.77               & 0.39           & 0.37           &   \textbf{0.16}        & 270 \\
        magistrale1             & 4.52          & \underline{2.19}   & 3.49           & \textbf{1.20}  &   2.35                 & 918 \\
        magistrale2             & 13.43         & 3.11               & 2.73           & \textbf{1.11}  &   \underline{2.24}     & 561 \\
        magistrale3             & 14.80         & \textbf{0.40}      & 1.22           & 0.74           &   1.69                 & 566 \\
        magistrale4             & 39.73         & 5.12               & \textbf{0.77}  & 1.58           &   \underline{1.02}     & 688 \\
        magistrale5             & 3.47          & 0.85               & 1.62           & \textbf{0.60}  &   \underline{0.73}     & 458 \\
        magistrale6             & X             & 2.29               & 3.91           & 3.23           &   \textbf{1.19}        & 771 \\
        outdoors1               & 101.95        & \textbf{74.96}     & X              & 255.04         & 123.24                 & 2656 \\
        outdoors2               & 21.67         & 133.46             & 73.86          & 64.61          &  \textbf{12.76}        & 1601 \\
        outdoors3               & 26.10         & 36.99              & 32.38          & 38.26          &   \textbf{8.92}        & 1531 \\
        outdoors4               & X             & 16.46              & 19.51          & 17.53          &  \textbf{15.25}        & 928 \\
        outdoors5               & 54.32         & 130.63             & 13.12          & 7.89           &   \textbf{7.16}        & 1168 \\
        outdoors6               & 149.14        & 133.60             & 96.51          & 65.50          &  \textbf{34.86}        & 2045 \\
        outdoors7               & 49.01         & 21.90              & 13.61          & \textbf{4.07}  &      \underline{5.00}  & 1748 \\
        outdoors8               & 36.03         & 83.36              & 16.31          & 13.53          &   \textbf{2.11}        & 986 \\
        room1                   & 0.16          & 0.07               & 0.06           & 0.09           &   \textbf{0.03}        & 146 \\
        room2                   & 0.33          & \textbf{0.07}      & 0.11           & \textbf{0.07}  &   0.13                 & 142 \\
        room3                   & 0.15          & 0.11               & \textbf{0.07}  & 0.13           &   \underline{0.09}     & 135 \\
        room4                   & 0.09          & \underline{0.04}   & \textbf{0.03}  & 0.05           &   \underline{0.04}     & 68 \\
        room5                   & 0.12          & 0.20               & 0.07           & 0.13           &   \textbf{0.06}        & 131 \\
        room6                   & 0.05          & 0.08               & 0.04           & \textbf{0.02}  &   \textbf{0.02}        & 67 \\
        slides1                 & 13.73         & 0.68               & 0.86           & 0.32           &   \textbf{0.31}        & 289 \\
        slides2                 & 0.81          & \underline{0.84}   & 2.15           & \textbf{0.32}  &   0.87                 & 299 \\
        slides3                 & 4.68          & 0.69               & 2.58           & 0.89           &    \textbf{0.60}       & 383 \\
        \textbf{avg drift}\%    & 16.83*        & 1.700              & 0.815*         & 0.939          & \textbf{0.472}         & normalized  \\ 

        \bottomrule
    \end{tabular}
}
\end{table}

The TUM-VI dataset~\cite{tumvi} is a very challenging handheld dataset, featuring large-scale indoor and outdoor scenes, and even sequences sliding down a tube, where almost the full image is covered. With long periods of walking in straight lines, stereo methods have an advantage here as they still can observe the scale with constant motion. 
We compare to the state-of-the-art visual-inertial odometry methods evaluated in~\cite{tumvi} in Table~\ref{tab:tumvi}.
Our method clearly outperforms the other monocular method in VINS-Mono~\cite{vinsmono} on most sequences, and even compared to the stereo methods it shows the best result on 16 sequences and a mean drift of 0.472. The closest competitor is again Basalt, which achieves the best result on 8 sequences and a mean drift of 0.939.

On this dataset, we also evaluate against ORB-SLAM3~\cite{orbslam3}, which is the state-of-the art visual-inertial SLAM system. This is not entirely fair as ORB-SLAM3 uses loop closures (which cannot be disabled), constituting an advantage over the other methods. We find the comparison still helpful as it allows to make conclusions regarding the underlying odometry.
We have evaluated ORB-SLAM3 5 times on each sequence and reproduced their results with code and settings provided by the authors.
For this comparison we have also evaluated VI-DSO~\cite{vidso}, and the results are shown in Fig.~\ref{fig:tumvi-cumulative}.
We observe that ORB-SLAM3 is more accurate on some sequences thanks to its very strong loop closure system. However, our method is more robust overall. This indicates that an integration of loop closure and map reuse into our system would be an interesting future research direction.

\begin{figure}[tb]
    \centering
    \includegraphics[width=\linewidth]{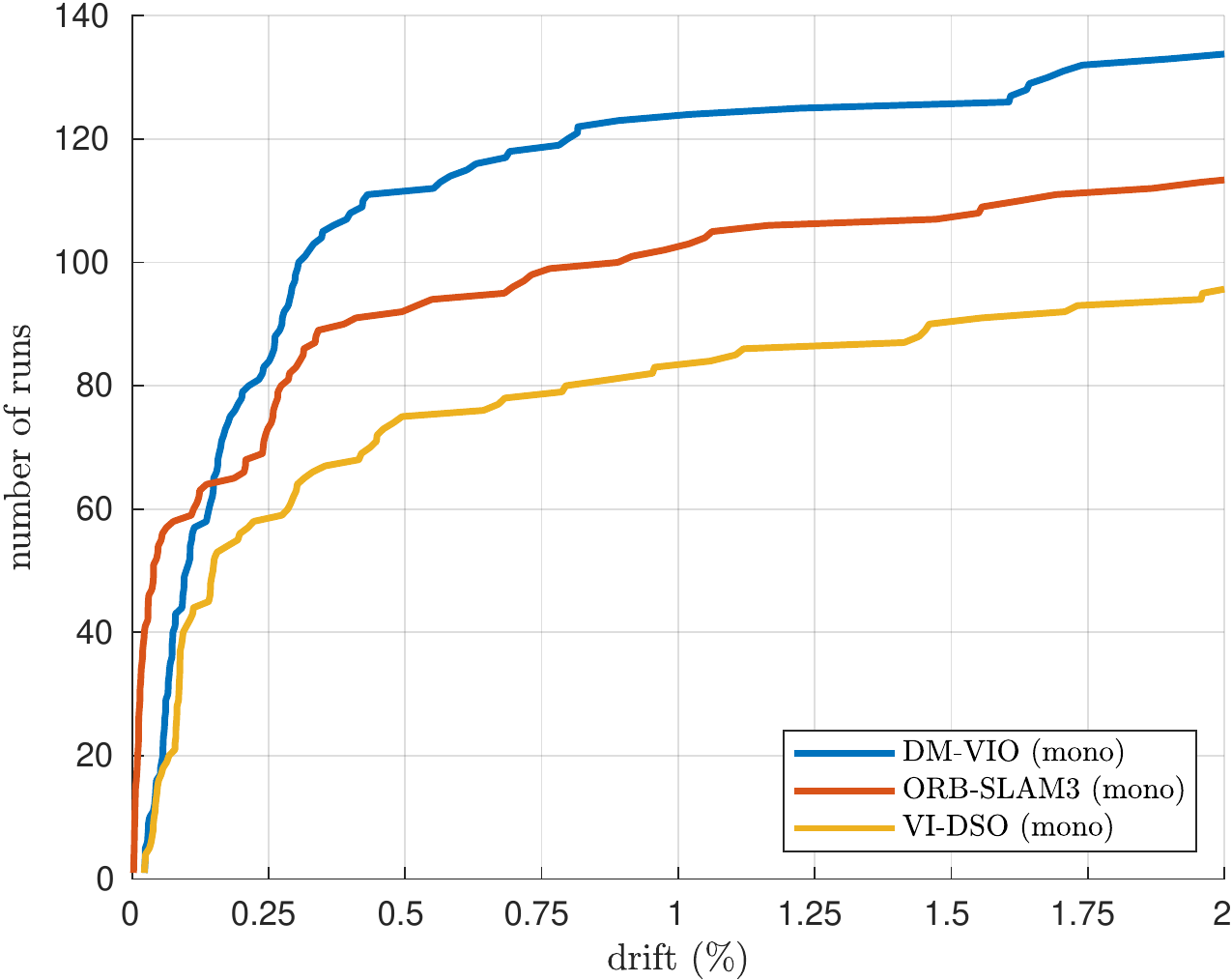} 
    \caption{Cumulative error plot for the TUM-VI dataset (drift in \%). Our method clearly outperforms both VI-DSO and ORB-SLAM3 in terms of robustness. Thanks to its powerful loop closure system, ORB-SLAM3 has an advantage in terms of accuracy on some sequences.}
    \label{fig:tumvi-cumulative} 
\end{figure}

\subsection{4Seasons dataset}

\begin{figure}[htb]
    \centering
    \includegraphics[width=\linewidth]{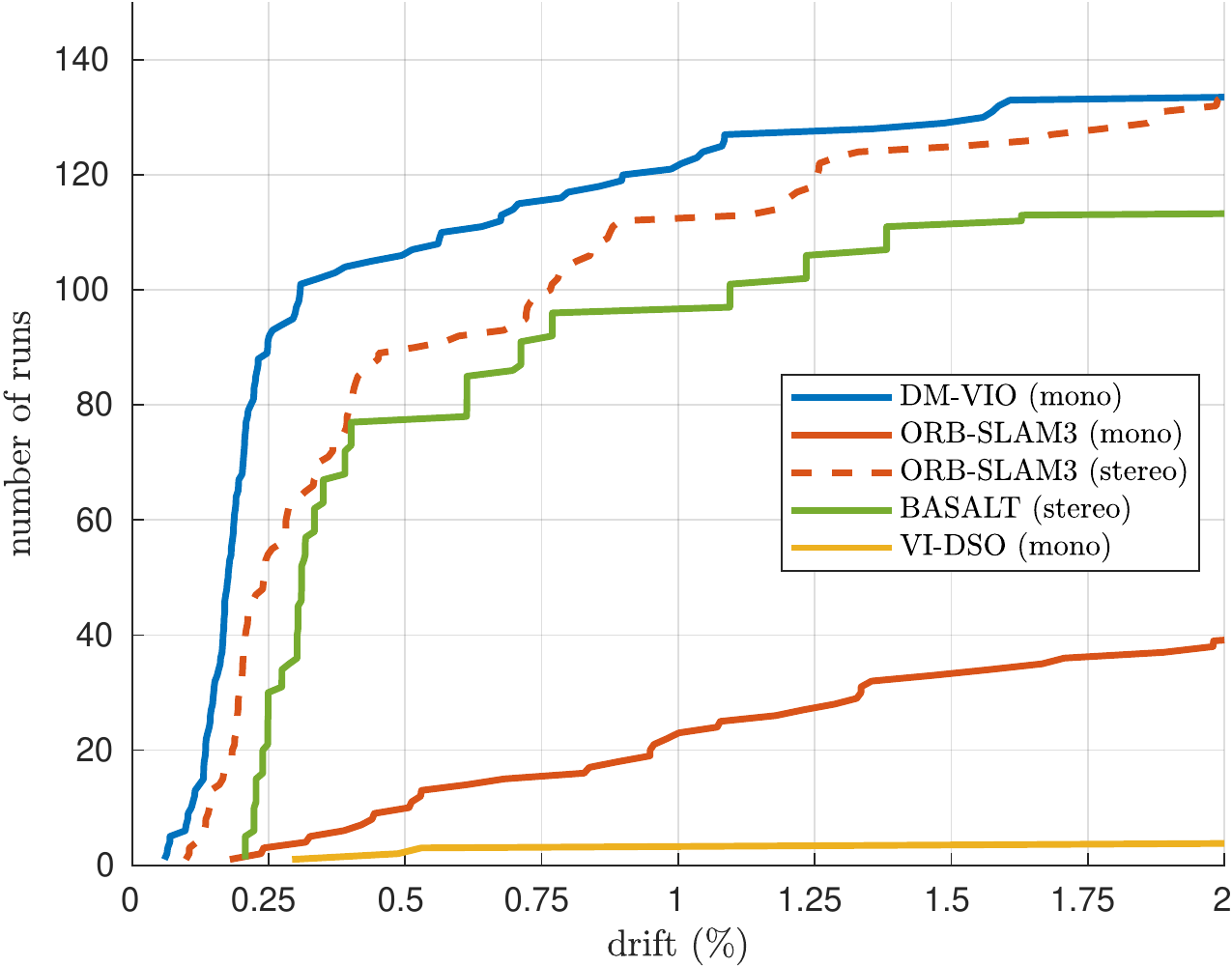} 
    \caption{Cumulative error plot for the 4Seasons dataset (drift in \%). With lots of stretches with constant velocity, this dataset is extremely challenging for monocular visual-inertial methods. Thanks to our novel IMU initializer powered by delayed marginalization and PGBA, DM-VIO is able to cope with it and even outperforms stereo-inertial methods.}
    \label{fig:4seasons-cumulative} 
\end{figure}

The 4Seasons dataset~\cite{4seasons} is a very recent automotive dataset, which, in contrast to most other car datasets, features a well time-synchronized visual-inertial sensor.
The lower part of the images is obstructed by the car hood, hence we crop off the bottom 96 pixels, which we do for all methods.
As this is the first odometry method to evaluate on the 4Seasons dataset, we make sure to determine IMU noise parameters for all methods the same way to ensure a fair comparison: We have manually read off the accelerometer and gyroscope noise density and bias random walk from the Allan variance plot provided in the data sheet of the IMU. To handle unmodeled effects we follow \cite{tumvi} and inflate noise values by different amounts to determine the best setting for all methods. For each method we tried noise models inflated by 1, 10, 100, 1000 respectively and chose the configuration which gave best results. 
For VI-DSO and for our method we slightly modified the visual initializer by adding a zero-prior to the translation on the x and y axis, and also added a threshold to stop keyframe creation for translations smaller than 0.01m (the latter was not activated for VI-DSO as it did not improve the results for it). Otherwise, parameters are the same as for the other experiments. 
For Basalt we tried all three provided default configurations with the optimal noise values to find the best settings.
After choosing the configuration for each method, we perform one final evaluation, running all 30 sequences 5 times each.

The results are shown in Fig.~\ref{fig:4seasons-cumulative}.
It is clear that the automotive scenario is very challenging for monocular methods. This is expected as it naturally features many stretches with constant motion, where scale is not observable, constituting a challenge for IMU initialization.
Thanks to our novel IMU initialization, DM-VIO not only works well on the dataset but even outperforms stereo-inertial ORB-SLAM3 and Basalt, while using monocular images and no loop closures.

\section{Conclusion and Future Work}
We have presented a monocular visual-inertial odometry system which outperforms the state of the art, even stereo-inertial methods.
Thanks to a novel IMU initializer, it works well in flying, handheld, and automotive scenarios, extending the applicability of monocular methods. 
The foundation of our IMU initialization is delayed marginalization, which also enables the pose graph bundle adjustment.

We anticipate that this method will spark further research in this direction.
The idea of delayed marginalization could be applied to more use cases, e.g. for reactivating old keyframes in a marginalization setting to enable map reuse.
The pose graph bundle adjustment can also be applied to long-term loop closures.
Lastly, our open-source system is easily extendible, as all optimizations 
are integrated with GTSAM, allowing to quickly add new factors.
This could be used for GPS integration, wheel odometry, and more.




\bibliographystyle{IEEEtran}
\bibliography{root}  

\begin{thebibliography}{10}
\providecommand{\url}[1]{#1}
\csname url@rmstyle\endcsname
\providecommand{\newblock}{\relax}
\providecommand{\bibinfo}[2]{#2}
\providecommand\BIBentrySTDinterwordspacing{\spaceskip=0pt\relax}
\providecommand\BIBentryALTinterwordstretchfactor{4}
\providecommand\BIBentryALTinterwordspacing{\spaceskip=\fontdimen2\font plus
\BIBentryALTinterwordstretchfactor\fontdimen3\font minus
  \fontdimen4\font\relax}
\providecommand\BIBforeignlanguage[2]{{%
\expandafter\ifx\csname l@#1\endcsname\relax
\typeout{** WARNING: IEEEtran.bst: No hyphenation pattern has been}%
\typeout{** loaded for the language `#1'. Using the pattern for}%
\typeout{** the default language instead.}%
\else
\language=\csname l@#1\endcsname
\fi
#2}}

\bibitem{msckf}
A.~I. Mourikis and S.~I. Roumeliotis, ``A multi-state constraint {K}alman
  filter for vision-aided inertial navigation,'' in \emph{ICRA}, 2007.

\bibitem{Kaiser14imu_init}
J.~Kaiser, A.~Martinelli, F.~Fontana, and D.~Scaramuzza, ``Simultaneous state
  initialization and gyroscope bias calibration in visual inertial aided
  navigation,'' \emph{RA-L}, vol.~2, no.~1, 2017.

\bibitem{vinsmono}
T.~Qin, P.~Li, and S.~Shen, ``{VINS-Mono: A} robust and versatile monocular
  visual-inertial state estimator,'' \emph{T-RO}, vol.~34, no.~4, 2018.

\bibitem{orbslamvi}
R.~Mur{-}Artal and J.~D. Tard{\'{o}}s, ``Visual-inertial monocular slam with
  map reuse,'' \emph{RA-L}, vol.~2, no.~2, 2017.

\bibitem{orbslam3}
C.~Campos, R.~Elvira, J.~J.~G. Rodríguez, J.~M. M.~Montiel, and
  J.~D.~Tard{\'{o}}s, ``{ORB-SLAM3: An} accurate open-source library for
  visual, visual-inertial, and multimap slam,'' \emph{T-RO}, pp. 1--17, 2021.

\bibitem{vidso}
L.~von Stumberg, V.~Usenko, and D.~Cremers, ``Direct sparse visual-inertial
  odometry using dynamic marginalization,'' in \emph{ICRA}, May 2018.

\bibitem{euroc}
M.~Burri, J.~Nikolic, P.~Gohl, T.~Schneider, J.~Rehder, S.~Omari, M.~W.
  Achtelik, and R.~Siegwart, ``The {EuRoC} micro aerial vehicle datasets,''
  \emph{IJRR}, 2016.

\bibitem{tumvi}
D.~Schubert, T.~Goll, N.~Demmel, V.~Usenko, J.~Stueckler, and D.~Cremers, ``The
  {TUM VI} benchmark for evaluating visual-inertial odometry,'' in \emph{IROS},
  October 2018.

\bibitem{4seasons}
P.~Wenzel, R.~Wang, N.~Yang, Q.~Cheng, Q.~Khan, L.~von Stumberg, N.~Zeller, and
  D.~Cremers, ``{4Seasons}: A cross-season dataset for multi-weather {SLAM} in
  autonomous driving,'' in \emph{GCPR}, 2020.

\bibitem{nister2004_vo}
D.~Nister, O.~Naroditsky, and J.~Bergen, ``Visual odometry,'' in \emph{CVPR},
  vol.~1, June 2004, pp. 652--659.

\bibitem{davison07pami}
A.~Davison, I.~Reid, N.~Molton, and O.~Stasse, ``{MonoSLAM}: Real-time single
  camera {SLAM},'' \emph{TPAMI}, vol.~29, 2007.

\bibitem{ptam}
G.~Klein and D.~Murray, ``Parallel tracking and mapping for small {AR}
  workspaces,'' in \emph{ISMAR}, 2007.

\bibitem{orbslam}
R.~Mur-Artal, J.~M.~M. Montiel, and J.~D. Tard{\'{o}}s, ``{ORB-SLAM: A}
  versatile and accurate monocular slam system,'' \emph{T-RO}, vol.~31, Oct
  2015.

\bibitem{kerl13icra}
C.~Kerl, J.~Sturm, and D.~Cremers, ``Robust odometry estimation for {RGB-D}
  cameras,'' in \emph{ICRA}, 2013.

\bibitem{dtam}
R.~Newcombe, S.~Lovegrove, and A.~Davison, ``{DTAM}: Dense tracking and mapping
  in real-time,'' in \emph{ICCV}, 2011.

\bibitem{lsdslam}
J.~Engel, T.~Sch\"ops, and D.~Cremers, ``{LSD-SLAM}: Large-scale direct
  monocular {SLAM},'' in \emph{ECCV}, 2014.

\bibitem{dso}
J.~Engel, V.~Koltun, and D.~Cremers, ``Direct sparse odometry,'' \emph{TPAMI},
  vol.~40, 2018.

\bibitem{rovio}
M.~Bloesch, S.~Omari, M.~Hutter, and R.~Siegwart, ``Robust visual inertial
  odometry using a direct {EKF}-based approach,'' in \emph{IROS}, 2015.

\bibitem{okvis}
S.~Leutenegger, S.~Lynen, M.~Bosse, R.~Siegwart, and P.~Furgale,
  ``Keyframe-based visual-inertial odometry using nonlinear optimization,''
  \emph{IJRR}, 2014.

\bibitem{basalt}
V.~Usenko, N.~Demmel, D.~Schubert, J.~Stueckler, and D.~Cremers,
  ``Visual-inertial mapping with non-linear factor recovery,'' \emph{RA-L},
  vol.~5, no.~2, pp. 422--429, 2020.

\bibitem{kimera}
A.~Rosinol, M.~Abate, Y.~Chang, and L.~Carlone, ``Kimera: an open-source
  library for real-time metric-semantic localization and mapping,'' in
  \emph{ICRA}, 2020.

\bibitem{Martinelli2014}
A.~Martinelli, ``Closed-form solution of visual-inertial structure from
  motion,'' \emph{IJCV}, vol. 106, no.~2, 2014.

\bibitem{onlinescale}
E.~Hong and J.~Lim, ``Visual-inertial odometry with robust initialization and
  online scale estimation,'' \emph{Sensors}, vol.~18, p. 4287, 12 2018.

\bibitem{preintegration-forster}
C.~Forster, L.~Carlone, F.~Dellaert, and D.~Scaramuzza, ``{IMU} preintegration
  on manifold for efficient visual-inertial maximum-a-posteriori estimation,''
  in \emph{RSS}, 2015.

\bibitem{atallcosts}
K.~MacTavish and T.~D. Barfoot, ``At all costs: A comparison of robust cost
  functions for camera correspondence outliers,'' in \emph{CRV}, 2015.

\bibitem{preintegration1}
T.~Lupton and S.~Sukkarieh, ``Visual-inertial-aided navigation for high-dynamic
  motion in built environments without initial conditions,'' \emph{T-RO},
  vol.~28, no.~1, pp. 61--76, 2012.

\bibitem{preintegration2}
L.~Carlone, Z.~Kira, C.~Beall, V.~Indelman, and F.~Dellaert, ``Eliminating
  conditionally independent sets in factor graphs: A unifying perspective based
  on smart factors,'' in \emph{ICRA}, 2014.

\bibitem{fej}
G.~Huang, A.~I. Mourikis, and S.~Roumeliotis, ``A first-estimates jacobian ekf
  for improving slam consistency,'' in \emph{ISER}, 2008.

\bibitem{gtsam}
F.~Daellert and Others, ``Gtsam,'' \url{https://gtsam.org}.

\bibitem{benchmarkpaper}
J.~Delmerico and D.~Scaramuzza, ``A benchmark comparison of monocular
  visual-inertial odometry algorithms for flying robots,'' in \emph{ICRA},
  2018, pp. 2502--2509.

\bibitem{camposinit}
C.~Campos, J.~Montiel, and J.~D. Tard{\'{o}}s, ``Inertial-only optimization for
  visual-inertial initialization,'' \emph{ICRA}, pp. 51--57, 2020.

\end{thebibliography}

\clearpage
\includepdf[pages=1]{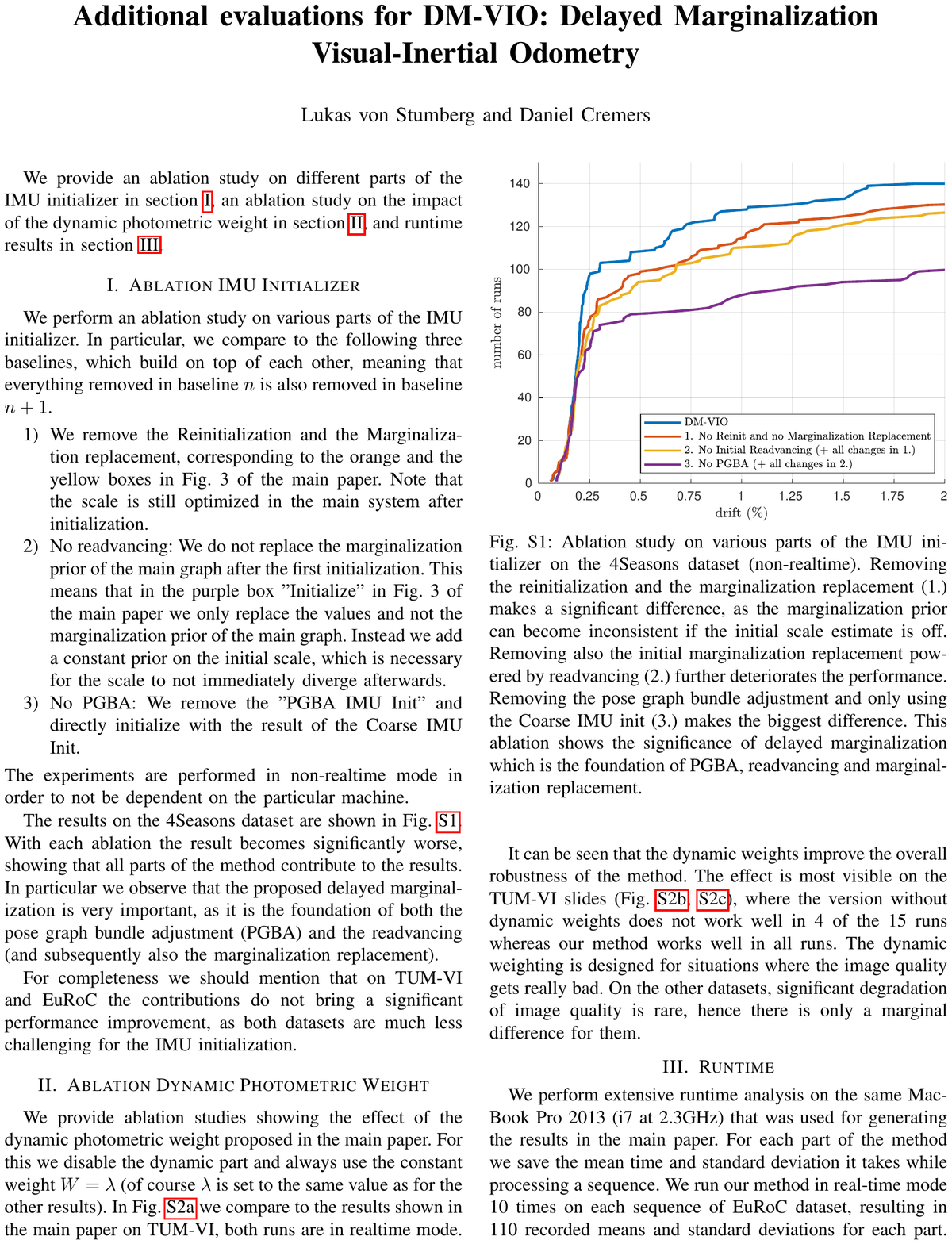}
\includepdf[pages=2]{supplementary.pdf}
\includepdf[pages=3]{supplementary.pdf}

\end{document}